\documentclass[letterpaper]{article} 
\usepackage{aaai2026}  
\usepackage{times}  
\usepackage{helvet}  
\usepackage{courier}  
\usepackage[hyphens]{url}  
\usepackage{graphicx} 
\usepackage{natbib}  
\usepackage{caption} 
\usepackage{algorithm}
\usepackage{algorithmic}
\usepackage{xcolor}

\usepackage{xurl}
\usepackage{newfloat}
\usepackage{listings}
\DeclareCaptionStyle{ruled}{labelfont=normalfont,labelsep=colon,strut=off} 
\floatstyle{ruled}
\newfloat{listing}{tb}{lst}{}
\floatname{listing}{Listing}
\usepackage{cleveref}
\crefname{equation}{eq.}{eqs.}
\Crefname{equation}{Eq.}{Eqs.}
\crefname{figure}{fig.}{figs.}
\Crefname{figure}{Fig.}{Figs.}
\crefname{table}{tab.}{tabs.}
\Crefname{table}{Tab.}{Tabs.}
\crefname{section}{sec.}{secs.}
\Crefname{section}{Sec.}{Secs.}
\usepackage{booktabs}
\usepackage{xspace}
\newcommand{\mymethod}{SMART\xspace}

\title{Analyzing Planner Design Trade-offs for MAPF under ADG-based Realistic Execution}
\author{
    Jingtian Yan\textsuperscript{\rm 1},
    Zhifei Li\textsuperscript{\rm 2},
    William Kang\textsuperscript{\rm 1},
    Stephen F. Smith\textsuperscript{\rm 1},
    Jiaoyang Li\textsuperscript{\rm 1}
}
\affiliations{
    \textsuperscript{\rm 1}Carnegie Mellon University,    
    \textsuperscript{\rm 2}Massachusetts Institute of Technology\\
    jingtianyan@cmu.edu, zhifeili@mit.edu, williamkang@cmu.edu, sfs@cs.cmu.edu, jiaoyangli@cmu.edu

}

\usepackage{bibentry}

\begin{document}

\maketitle

\begin{abstract}
Multi-Agent Path Finding (MAPF) algorithms are increasingly deployed in industrial warehouses and automated manufacturing facilities, where robots must operate reliably under real-world physical constraints. However, existing MAPF evaluation frameworks typically rely on simplified robot models, leaving a substantial gap between algorithmic benchmarks and practical performance.
Recent frameworks such as SMART combine kinodynamic modeling with execution based on the Action Dependency Graph (ADG), enabling realistic, large-scale MAPF evaluation.
Building on this capability, this work investigates how key planner design choices influence performance under realistic execution settings. We systematically study three fundamental factors: (1) the relationship between solution optimality and execution performance, (2) the sensitivity of system performance to inaccuracies in kinodynamic modeling, and (3) the tradeoff between model accuracy and plan optimality.
Empirically, we examine these factors to understand how these design choices affect performance in realistic scenarios.
We highlight open challenges and research directions to steer the community toward practical, real-world deployment.

\end{abstract}

\section{Introduction}
Multi-Agent Path Finding (MAPF) has emerged as a critical challenge for coordinating robot fleets in automated warehouses, manufacturing facilities, and logistics centers~\cite{ma2017feasibility,honig2019warehouse,Ho2022uav}.
As these systems continue to grow in scale and complexity, the demand for reliable and high-performance MAPF planners has intensified. Despite significant progress in algorithm design, a persistent challenge remains: most existing MAPF evaluation frameworks rely on overly simplified robot models, neglecting realistic kinodynamic constraints, execution-time variability, and other physical characteristics inherent to robotics systems. This gap limits the ability of researchers and practitioners to accurately assess how planner design choices translate into performance in realistic settings.

Some evaluation tools~\cite{heuer2024benchmarking, yan2025advancing} are proposed to evaluate the performance of MAPF methods in realistic settings.
However, only limited work has studied how planner design choices translate into execution performance in realistic scenarios. To the best of our knowledge, \citet{varambally2022mapf} present the only study in this direction. Their work compares different MAPF models used during planning in automated warehousing environments, providing valuable initial insights.
Nevertheless, their evaluation is limited to less than 50 robots, focuses solely on a warehouse layout, and does not systematically analyze how different planner design factors interact and collectively affect execution outcomes.

In this work, we investigate how key planner design choices influence MAPF performance under realistic settings using \mymethod~\cite{yan2025advancing}, a state-of-the-art testbed.
Built on the Action Dependency Graph (ADG)~\cite{honig2019warehouse}, \mymethod is a physics-based simulation framework that enables execution-aware evaluation of MAPF planners in realistic settings and supports large-scale experiments with thousands of robots.
Leveraging this capability, we systematically study the impact of three fundamental factors on execution performance:
(1) the plan optimality,
(2) the accuracy of the kinodynamic model used during planning, and
(3) the trade-off between model accuracy and plan optimality.
Through controlled experiments in \mymethod, we show how these factors influence execution behavior in practice.
Our analysis uncovers key planner design trade-offs and outlines open challenges and research directions for advancing MAPF toward real-world deployment.

\section{Background}
In this section, we first introduce the MAPF problem and related work. Next, we discuss existing evaluation tools.

\subsection{MAPF Problem}
A \emph{MAPF problem}~\cite{Stern2019benchmark} is typically defined on a discrete undirected graph $\mathcal{G}_M = (\mathcal{V}_M, \mathcal{E}_M)$, with a set of robots $\mathcal{R} = \{r_1, \dots, r_I\}$ where each robot has a start and a goal location. A \emph{MAPF plan} consists of collision-free paths for all robots to reach their goals. In the standard formulation, time is discretized into timesteps, and each robot can either move to an adjacent vertex or wait. Collisions occur if two robots occupy the same vertex or traverse the same edge in opposite directions at the same timestep. 

\subsection{MAPF Planners and Design Choices}

Many MAPF methods have been proposed.
Some methods focus on finding solutions with optimal or bounded-suboptimal guarantees~\cite{sharon2015conflict,barer2014suboptimal,li2021eecbs}, while others focus on scalability, solving MAPF problems involving thousands of robots~\cite{okumura2022priority}.
Anytime algorithms quickly find an initial solution and then iteratively refine it as time progresses, balancing solution quality with computational efficiency~\cite{li2021anytime,HuangAAAI22}.
Additionally, some methods incorporate complex real-world factors, such as kinodynamics and unexpected delays, into the planning process~\cite{cohen2019optimal,atzmon2020robust,yan2024PSB}.
Despite these advances, it remains unclear how such planner design decisions actually impact performance in realistic execution settings.
\citet{varambally2022mapf} investigate how different assumptions in MAPF modeling affect practical performance in automated warehousing.
However, its evaluation explores only a small set of MAPF modeling assumptions, focuses solely on warehouse layouts, and is limited to less than 50 robots.

\subsection{Evaluating MAPF in Realistic Scenarios}
Evaluating MAPF performance requires executing planned paths in settings that reflect real robotic behavior. Some tools have been developed to enable such assessment. MRP-Bench~\cite{schaefer2023benchmark} utilizes Gazebo with integrated low-level controllers to account for kinodynamics and motion delays.
The recently introduced SMART testbed~\cite{yan2025advancing} explicitly models robot kinodynamics, execution uncertainties, and communication delays.
SMART scales to thousands of robots, making it well suited for evaluating modern MAPF planners.

The architecture of SMART integrates a physics-engine-based simulator, an execution monitoring server based on the Action Dependency Graph (ADG)~\cite{honig2019warehouse}, and robot-specific executors.
Given a MAPF plan, SMART first uses the execution monitoring server to convert the plan into an ADG, a directed graph consisting of two types of edges: (i) Type-1 (intra-robot) edges, which encode the temporal order of actions executed by the same robot, and (ii) Type-2 (inter-robot) edges, which encode precedence constraints between actions of different robots to prevent collisions.
Then, the robot executors follow the ADG to issue motion commands to each robot while accounting for kinematic limits, delays, and disturbances. Finally, the simulator executes these commands in a physics-based simulation and returns realistic execution outcomes for evaluation.

\section{From MAPF to Realistic Scenarios}
In this section, we study research problems focusing on the gap between MAPF planning and robot execution.
We use six simulation environments derived from the MovingAI benchmark~\cite{Stern2019benchmark} maps:
\texttt{empty} (empty-32-32, size: 32$\times$32), \texttt{random} (random-64-64-10, size: 64$\times$64), \texttt{room} (room-64-64-16, size: 64$\times$64), \texttt{den312d} (den312d, size: 65$\times$81), \texttt{maze} (maze-32-32-4, size: 32$\times$32), and \texttt{warehouse} (warehouse-10-20-10-2-1, size: 161$\times$63).
We use the Average Execution Time (AET) to measure the execution performance of a MAPF plan, which is the total simulated time that all robots take to complete their assigned paths divided by the number of robots.
We adopt the same robot configuration as used in the \mymethod paper: all robots are modeled as differential-drive platforms with circular footprints of 0.35 m in diameter and are driven by PID controllers for action execution. They can move forward and perform in-place rotations, with translational and rotational velocities constrained to [0, 0.5] m/s and 30$^\circ$/s, respectively, and acceleration limited to [$-$0.4, 0.4] m/s\textsuperscript{2}.
We leave generalization across different robot kinodynamics to future work.
The grid cell is standardized to 1.0$\times$1.0 m.
All experiments are conducted on an Ubuntu 20.04 server with a 64-core AMD 7980X CPU and 256 GB RAM.

\begin{table*}[ht!]
\small
\centering
\resizebox{\linewidth}{!}{
\begin{tabular}{@{}lcccccccccc@{}}
\toprule
\textbf{Feature} 
& All Features 
& SoC + Rotations
& SoC + Conflict pairs 
& Rotations + Conflict pairs 
& SoC 
& Type-1 edges 
& Type-2 edges 
& Rotations 
& Conflict pairs 
& Robots \\ 
\midrule
\textbf{MAPE} 
& \textbf{0.0342} 
& 0.070
& 0.162
& 0.164
& 0.1778 
& 0.2414 
& 0.2826 
& 0.1793 
& 0.3175 
& 0.4317 \\
\bottomrule
\end{tabular}
}
\caption{Quadratic regression results. ``All Features'' uses the full feature set, the next three columns use two-feature combinations, and the remaining columns use single features.}
\label{tab:regression-results}
\end{table*}

\subsection{Impact of Optimality} \label{method:exp1}
We first study the practical effectiveness of the objective used by MAPF algorithms.
One of the most widely used objectives in MAPF research is the \emph{Sum of Costs} (SoC), which measures the total number of timesteps for all robots to reach their goals.
While extensive efforts have been made to develop algorithms that optimize SoC optimally or sub-optimally, it is unclear, for example, whether reducing SoC by 1\% translates to a similar degree of improvement in AET in a realistic simulation.
In this section, we assess the correlation between improvements in SoC and AET.
We also explore the possibility that there might be better objectives than SoC for evaluating the MAPF plan in realistic settings.

\subsubsection{Experiment Setup}
We generate MAPF plans with varying SoC for a MAPF instance by using MAPF-LNS\footnote{Code at \url{https://github.com/Jiaoyang-Li/MAPF-LNS}}~\cite{li2021anytime}, an anytime method that iteratively refines MAPF plans to improve SoC within a predefined runtime limit.
For each map, we select three different numbers of robots and five sets of different start and goal locations for robots.
For each MAPF instance, we record all plans generated in each iteration of MAPF-LNS within the runtime limit of 60 seconds.
We then execute these plans in \mymethod.

\begin{figure}
    \centering
    \includegraphics[width=\linewidth]{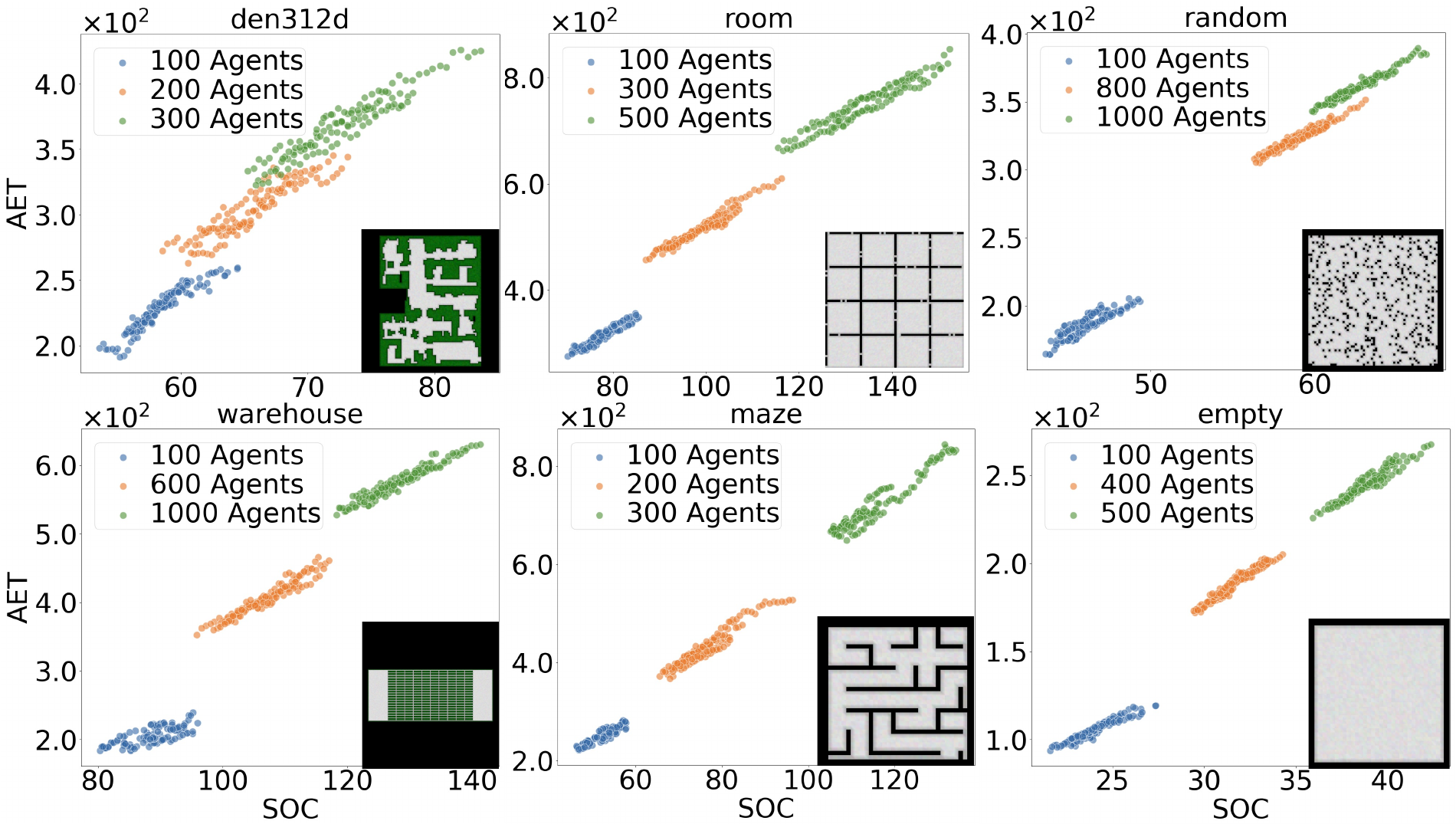}
    \caption{Relationship between SoC and AET. Each dot represents the SoC and AET of a single MAPF plan.}
    \label{fig:exp1_soc_AET}
\end{figure}

\subsubsection{Results and Analysis}
We first study how well the SoC objective measures the execution time of robots.
As shown in~\cref{fig:exp1_soc_AET}, SoC and AET show a strong positive linear correlation across all maps and all numbers of robots, indicating that SoC captures the trend of AET.
However, the correlation is not perfectly monotonic.
For the same SoC value, there are noticeable variations in AET, suggesting that SoC alone cannot account for all factors influencing execution time.

To explore other factors that influence execution time, we extract additional features from the ADG, including (1) \emph{Type-1 edges}, the number of type-1 edges, (2) \emph{Type-2 edges}, the number of type-2 edges, (3) \emph{Rotations}, the number of rotation actions, (4) \emph{Conflict pairs}, the number of robot pairs connected by type-2 edges, and (5) \emph{Robots}, the number of robots.
We analyze the relationship between these features and AET using a quadratic regression model and measure prediction accuracy with mean absolute percentage error (MAPE). As shown in~\cref{tab:regression-results}, SoC has the strongest predictive power among individual features, while other features also show correlation with AET.
Among the tested feature pairs, \emph{SoC + Rotations} performs best.
Notably, the model using all features achieves a much lower MAPE than any single feature or feature combination, indicating that execution time is influenced by multiple factors.

\begin{figure}
    \centering
    \includegraphics[width=\linewidth]{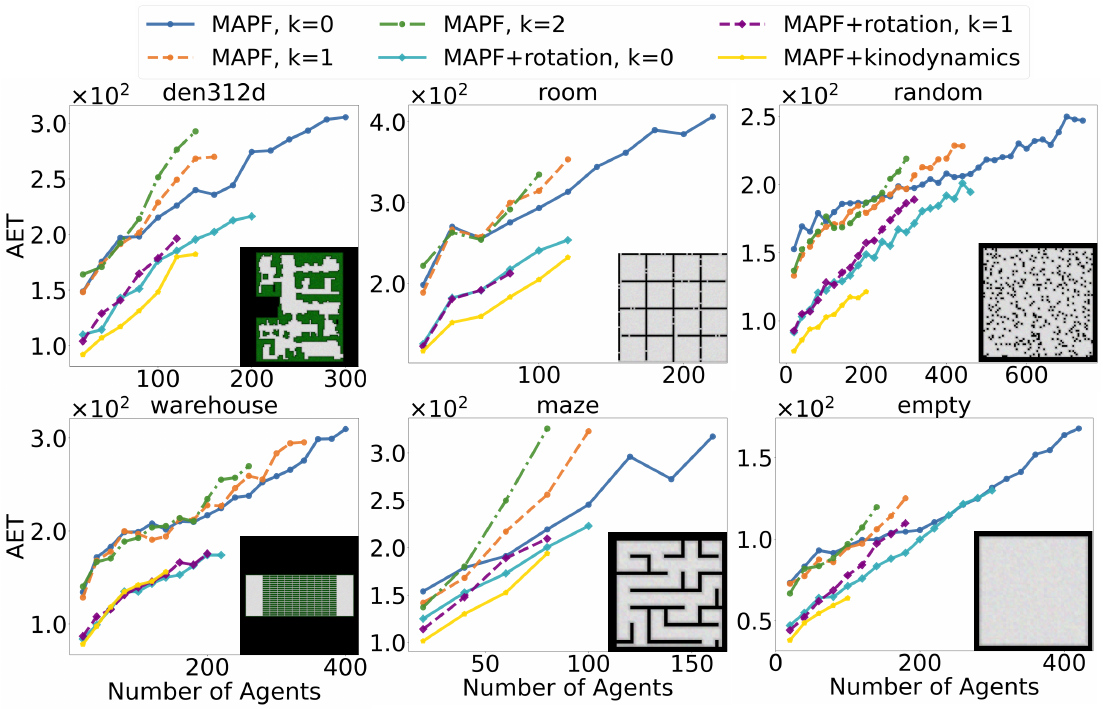}
    \caption{AET using different MAPF models.}
    \label{fig:exp2_sol_compare}
\end{figure}

\begin{figure}
    \centering
    \includegraphics[width=\linewidth]{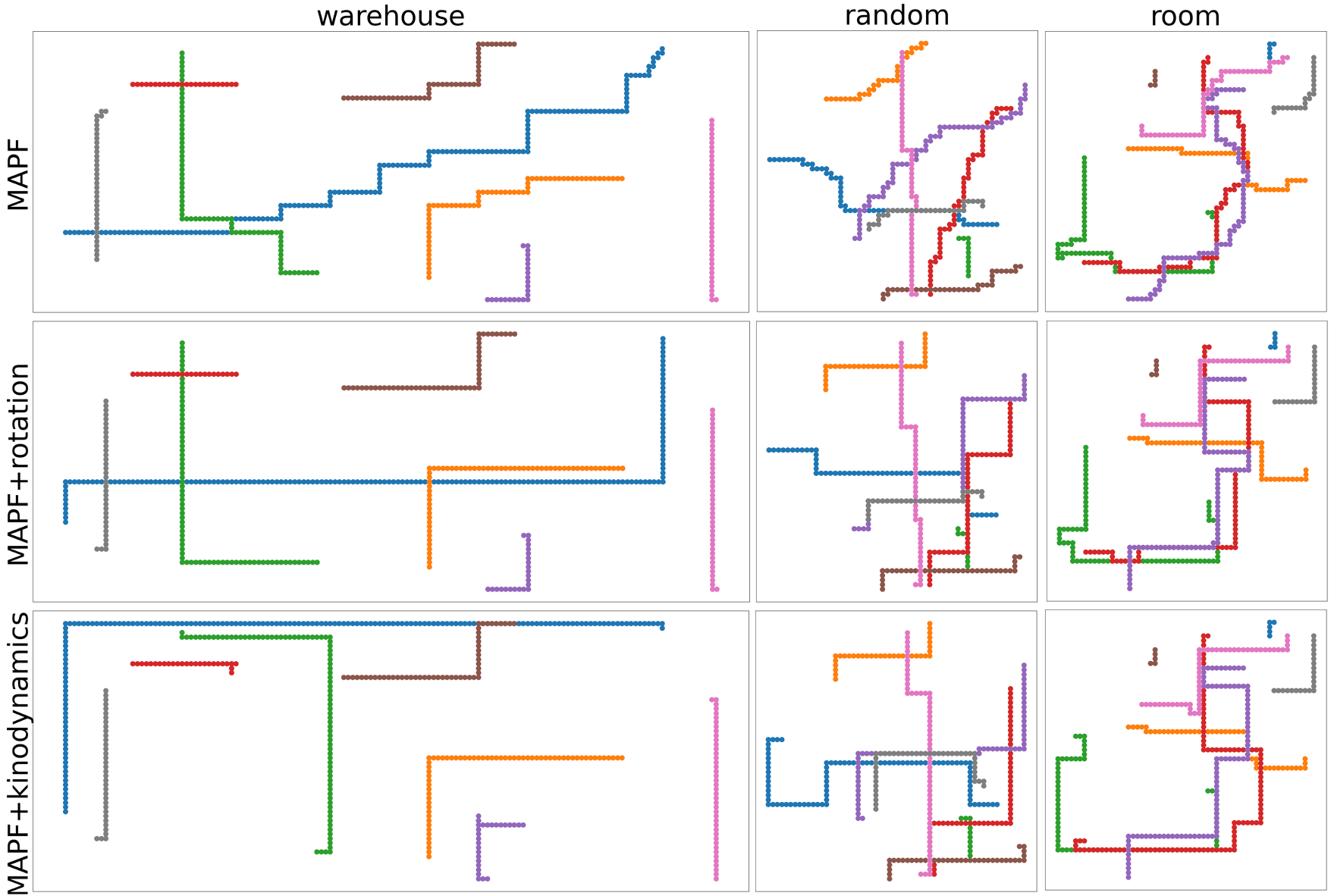}
    \caption{Paths generated on different maps.}
    \label{fig:exp2_paths}
\end{figure}

\subsection{Impact of Different MAPF models} \label{method:exp2}
The standard MAPF problem uses a simplified model that ignores execution factors such as kinodynamics and potential delays, creating a gap between planning and execution. Thus, recent research has introduced more realistic MAPF models.
An ideal MAPF model should capture the essential aspects of robot kinodynamics while omitting unnecessary details. Thus, we conduct an empirical comparison of different MAPF models to identify the most effective one.

\subsubsection{Experiment Setup}
We consider the following MAPF models: (1) the standard MAPF model, (2) the MAPF model with rotation~\cite{zhang2023efficient}, and (3) the MAPF model that accounts for the full kinodynamics of robots, including rotation, speed, and acceleration~\cite{yan2025multi}.
To account for execution delays, we apply the $k$-robust delay model~\cite{atzmon2020robust}. This model introduces bounded temporal slack of up to $k$ time steps during planning to tolerate potential execution delays. Combining these factors produces three variants: \emph{MAPF, $k=i$}, \emph{MAPF+rotation, $k=i$}, and \emph{MAPF+kinodynamics}.
We use PBS\footnote{Code at \url{https://github.com/Jiaoyang-Li/PBS}}~\cite{ma2019searching} to handle conflicts between robots with a runtime limit of 60 seconds.
We run these methods to generate MAPF plans for all maps using 25 ``random scenarios'' from the MAPF benchmark~\cite{Stern2019benchmark} with a progressive number of robots until the success rate is lower than 20\%.
Finally, these plans are executed in \mymethod.

\subsubsection{Results and Analysis}
As shown in~\cref{fig:exp2_sol_compare}, the method incorporating the full kinodynamics of robots achieves the best AET among all the models. Additionally, the planner considering robot rotations demonstrates much better AET than the standard MAPF model.
These results indicate that incorporating more accurate MAPF models into planning can improve execution time.
However, we also observe trade-offs associated with these models.
Incorporating more accurate MAPF models often reduces scalability. For example, modeling rotations during planning can reduce the maximum number of solvable agents by up to 40\%.
Meanwhile, the $k$-robust model performs comparably across different $k$ values with fewer agents; however, its performance declines significantly as the number of robots increases, resulting in lower solution quality and reduced scalability.
Interestingly, in the \texttt{warehouse} map, MAPF+kinodynamics does not provide significant improvements over MAPF+rotation. To better understand this, we visualize the MAPF plans in~\cref{fig:exp2_paths}. 
The results show that the paths in \texttt{warehouse} have fewer shared vertices than in the other two maps.
We hypothesize that this minimizes the potential benefits of accounting for kinodynamics.
In summary, while more accurate MAPF models generally improve execution time in realistic settings, their increased computational complexity can limit scalability. Additionally, the $k$-robust model is less efficient and may even harm performance, particularly in more congested scenarios.

\subsection{Optimality and MAPF Models}
From \cref{method:exp1,method:exp2}, we show that both improving SoC and incorporating a more accurate MAPF model during planning can lead to better AET during execution.
However, both of them require longer computation time.
Considering the limited planning time available in real-world applications, this section explores the balance between these factors to provide practical insights.

\subsubsection{Experiment Setup}
We generate MAPF plans with varying SoC for three MAPF models: (1) standard MAPF, (2) MAPF with rotation, and (3) MAPF with kinodynamics.
To obtain optimal solutions, we use CBS\footnote{Code at \url{https://github.com/Jiaoyang-Li/CBSH2-RTC}}~\cite{li2021pairwise} for the standard MAPF model and CBS with rotation\footnote{Code at \url{https://github.com/YueZhang-studyuse/MAPF_T}}~\cite{zhang2023efficient} for the rotation model.
For suboptimal plans, we combine PBS with all three models, as described in~\cref{method:exp2}.
To analyze the relationship between AET and SoC, we create a naive anytime planner by running Prioritized Planning (PP)~\cite{erdmann1987multiple} with random restarts and recording the MAPF plan with the lowest SoC until reaching a runtime limit of 60 seconds.
We do not use existing anytime solvers, such as MAPF-LNS, because, to our best knowledge, no available implementation considers MAPF models with kinodynamic constraints. 

\begin{figure}
    \centering
    \includegraphics[width=\linewidth]{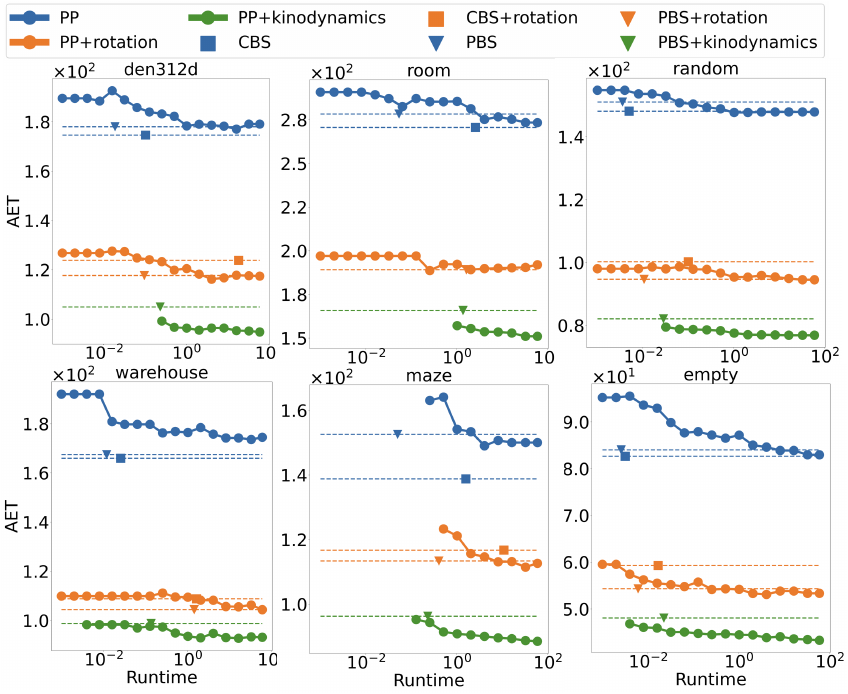}
    \caption{AET using different MAPF models and planners.}
    \label{fig:exp3_sol_model}
\end{figure}

\subsubsection{Results and Analysis}
As shown in \cref{fig:exp3_sol_model}, planners that use more accurate MAPF models consistently achieve better AET, even when less accurate models produce optimal solutions.
Concretely, adding rotation reduces AET by 27--33\% compared to the standard MAPF model, and adding full kinodynamic constraints provides an additional 17--20\% improvement.
This indicates that incorporating more accurate MAPF models into planning may be more important than obtaining optimal solutions with less accurate models.
Notably, as discussed in~\cref{method:exp1}, since SoC is not a perfect metric, the optimal planner does not always achieve the best AET.
Meanwhile, execution frameworks may also become a bottleneck. Since ADG adopts a conservative strategy, assuming robots can be delayed indefinitely, it compromises the solution quality of the MAPF plan in certain cases.
Overall, achieving better AET depends on both more accurate MAPF models and more effective execution frameworks.

\section{Challenges and Future Directions}
Our study highlights several challenges that motivate future directions for the deployment of MAPF in realistic settings.
\paragraph{Using Execution-Aware Objectives}
Although SoC, the most widely used objective, is a good approximation of execution performance, our results show that incorporating additional execution-related features can yield more accurate estimates of execution time. Thus, designing execution-aware objective functions and MAPF planners that optimize them represents an important research opportunity.

\paragraph{Developing Efficient Planners for More Accurate MAPF Models}
Our results show that using even slightly more accurate MAPF models during planning can significantly improve execution performance. However, this comes at the cost of substantially reduced scalability. Over the past decade, the MAPF community has made remarkable progress in scaling up planners for the standard MAPF model, yet planners that consider kinodynamics still struggle to scale. This underscores the need for better MAPF models and more efficient planners to solve them while maintaining the scalability required for industrial deployment.

\paragraph{Balancing Planner Optimality and Model Accuracy}
When computation time is limited, our results show that model accuracy often improves execution performance more than aggressive refinement of SoC. This highlights an underexplored trade-off between planner optimality and model accuracy. Moreover, it suggests the need for planners that can adaptively balance these factors based on time budgets.

\paragraph{Enhancing Execution Frameworks}
Another direction is to develop execution frameworks that operate less conservatively than ADG.
Even when planning with accurate robot models, execution errors, like communication delays and controller variability, can accumulate and degrade execution performance.
Recent methods, such as Switchable TPG~\cite{berndt2023receding,JiangAAAI25} and Bidirectional TPG~\cite{su2024bidirectional}, dynamically adjust passing orders between robots to mitigate such execution errors. Adapting these ideas to our execution framework and further advancing this direction could enhance execution performance while maintaining robustness in large-scale industrial deployments.

\newpage

\section*{Acknowledgments}
This work is in part supported by the National Science Foundation (NSF) under grant numbers \#$2328671$ and \#$2441629$, as well as a gift from Amazon.

\bibliography{icaps}

@string{AAAI = {Proceedings of the AAAI Conference on Artificial Intelligence}}

@string{ICRA = {Proceedings of the IEEE International Conference on Robotics and Automation}}

@string{SoCS = {Proceedings of the International Symposium on Combinatorial Search}}

@string{IJCAI = {Proceedings of the International Joint Conference on Artificial Intelligence}}

@string{RAL = {IEEE Robotics and Automation Letters}}

@inproceedings{su2024bidirectional,
  title={Bidirectional Temporal Plan Graph: {E}nabling Switchable Passing Orders for More Efficient Multi-Agent Path Finding Plan Execution},
  author={Su, Yifan and Veerapaneni, Rishi and Li, Jiaoyang},
  booktitle=AAAI,
  volume={38},
  pages={17559--17566},
  year={2024}
}

@article{yan2025advancing,
  title={Advancing {MAPF} Towards the Real World: {A} Scalable Multi-Agent Realistic Testbed ({SMART})},
  author={Yan, Jingtian and Li, Zhifei and Kang, William and Zheng, Kevin and Zhang, Yulun and Chen, Zhe and Zhang, Yue and Harabor, Daniel and Smith, Stephen F and Li, Jiaoyang},
  journal={arXiv preprint arXiv:2503.04798},
  year={2025}
}

@inproceedings{ma2019searching,
  title={Searching with Consistent Prioritization for Multi-Agent Path Finding},
  author={Ma, Hang and Harabor, Daniel and Stuckey, Peter J and Li, Jiaoyang and Koenig, Sven},
  booktitle=AAAI,
  volume={33},
  pages={7643--7650},
  year={2019}
}

@inproceedings{cohen2019optimal,
  title={Optimal and Bounded-Suboptimal Multi-Agent Motion Planning},
  author={Cohen, Liron and Uras, Tansel and Kumar, T. K. Satish and Koenig, Sven},
  booktitle=SoCS,
  volume={10},
  pages={44--51},
  year={2019}
}

@inproceedings{Stern2019benchmark,
  author    = {Roni Stern and Nathan R. Sturtevant and Ariel Felner and Sven Koenig and Hang Ma and Thayne T. Walker and Jiaoyang Li and Dor Atzmon and Liron Cohen and T. K. Satish Kumar and Eli Boyarski and Roman Bart{\'{a}}k},
  title     = {Multi-Agent Pathfinding: Definitions, Variants, and Benchmarks},
  booktitle = SoCS,
  pages     = {151--159},
  year      = {2019}
}

@inproceedings{varambally2022mapf,
  title={Which {MAPF} Model Works Best for Automated Warehousing?},
  author={Varambally, Sumanth and Li, Jiaoyang and Koenig, Sven},
  booktitle=SoCS,
  volume={15},
  pages={190--198},
  year={2022}
}

@ARTICLE{honig2019warehouse,
  author={H\"{o}nig, Wolfgang and Kiesel, Scott and Tinka, Andrew and Durham, Joseph W. and Ayanian, Nora},
  journal=RAL, 
  title={Persistent and Robust Execution of {MAPF} Schedules in Warehouses}, 
  year={2019},
  volume={4},
  number={2},
  pages={1125--1131}}

@inproceedings{li2021eecbs,
  title={{EECBS}: A Bounded-Suboptimal Search for Multi-Agent Path Finding},
  author={Li, Jiaoyang and Ruml, Wheeler and Koenig, Sven},
  booktitle=AAAI,
  volume={35},
  pages={12353--12362},
  year={2021}
}

@article{li2021pairwise,
  title={Pairwise Symmetry Reasoning for Multi-Agent Path Finding Search},
  author={Li, Jiaoyang and Harabor, Daniel and Stuckey, Peter J and Ma, Hang and Gange, Graeme and Koenig, Sven},
  journal={Artificial Intelligence},
  volume={301},
  pages={103574},
  year={2021}
}

@article{sharon2015conflict,
  title={Conflict-Based Search For Optimal Multi-Agent Path Finding},
  author={Sharon, Guni and Stern, Roni and Felner, Ariel and Sturtevant, Nathan R},
  journal={Artificial Intelligence},
  volume={219},
  pages={40--66},
  year={2015}
}

@ARTICLE{yan2024PSB,
  author={Yan, Jingtian and Li, Jiaoyang},
  journal=RAL, 
  title={Multi-Agent Motion Planning With B\'ezier Curve Optimization Under Kinodynamic Constraints}, 
  year={2024},
  volume={9},
  number={3},
  pages={3021--3028}}

@article{erdmann1987multiple,
  title={On Multiple Moving Objects},
  author={Erdmann, Michael and Lozano-Perez, Tomas},
  journal={Algorithmica},
  volume={2},
  pages={477--521},
  year={1987}
}

@article{okumura2022priority,
  title={Priority Inheritance with Backtracking for Iterative Multi-agent Path Finding},
  author={Okumura, Keisuke and Machida, Manao and D{\'e}fago, Xavier and Tamura, Yasumasa},
  journal={Artificial Intelligence},
  volume={310},
  pages={103752},
  year={2022}
}

@article{atzmon2020robust,
  title={Robust Multi-Agent Path Finding and Executing},
  author={Atzmon, Dor and Stern, Roni and Felner, Ariel and Wagner, Glenn and Bart{\'a}k, Roman and Zhou, Neng-Fa},
  journal={Journal of Artificial Intelligence Research},
  volume={67},
  pages={549--579},
  year={2020}
}

@inproceedings{zhang2023efficient,
  title={Efficient Multi Agent Path Finding with Turn Actions},
  author={Zhang, Yue and Harabor, Daniel and Le Bodic, Pierre and Stuckey, Peter J},
  booktitle=SOCS,
  volume={16},
  pages={119--127},
  year={2023}
}

@inproceedings{JiangAAAI25,
  author    = {He Jiang and Muhan Lin and Jiaoyang Li},
  title     = {Speedup Techniques For Switchable {T}emporal {P}lan {G}raph Optimization},
  booktitle = AAAI,
  pages     = {23212--23221},
  year      = {2025}
}

@inproceedings{yan2025multi,
  title={Multi-Agent Motion Planning For Differential Drive Robots Through Stationary State Search},
  author={Yan, Jingtian and Li, Jiaoyang},
  booktitle=AAAI,
  volume={39},
  pages={23360--23368},
  year={2025}
}

@inproceedings{heuer2024benchmarking,
  title={Benchmarking Multi-Robot Coordination in Realistic, Unstructured Human-Shared Environments},
  author={Heuer, Lukas and Palmieri, Luigi and Mannucci, Anna and Koenig, Sven and Magnusson, Martin},
  booktitle=ICRA,
  pages={14541--14547},
  year={2024}
}

@inproceedings{schaefer2023benchmark,
  title={A Benchmark for Multi-Robot Planning in Realistic, Complex and Cluttered Environments},
  author={Schaefer, Simon and Palmieri, Luigi and Heuer, Lukas and Dillmann, Ruediger and Koenig, Sven and Kleiner, Alexander},
  booktitle=ICRA,
  pages={9231--9237},
  year={2023}
}

@ARTICLE{Ho2022uav,
  author={Ho, Florence and Gonçalves, Artur and Rigault, Bastien and Geraldes, Rúben and Chicharo, Alexandre and Cavazza, Marc and Prendinger, Helmut},
  journal={IEEE Intelligent Transportation Systems Magazine}, 
  title={Multi-Agent Path Finding in Unmanned Aircraft System Traffic Management With Scheduling and Speed Variation}, 
  year={2022},
  volume={14},
  number={5},
  pages={8--21}}

@inproceedings{ma2017feasibility,
  title={Feasibility Study: {M}oving Non-Homogeneous Teams in Congested Video Game Environments},
  author={Ma, Hang and Yang, Jingxing and Cohen, Liron and Kumar, TK and Koenig, Sven},
  booktitle={Proceedings of the AAAI Conference on Artificial Intelligence and Interactive Digital Entertainment},
  volume={13},
  pages={270--272},
  year={2017}
}

@inproceedings{barer2014suboptimal,
  title={Suboptimal Variants of the Conflict-Based Search Algorithm for the Multi-Agent Pathfinding Problem},
  author={Barer, Max and Sharon, Guni and Stern, Roni and Felner, Ariel},
  booktitle=SOCS,
  volume={5},
  pages={19--27},
  year={2014}
}

@inproceedings{li2021anytime,
  title={Anytime Multi-Agent Path Finding via Large Neighborhood Search},
  author={Li, Jiaoyang and Chen, Zhe and Harabor, Daniel and Stuckey, Peter J and Koenig, Sven},
  booktitle=IJCAI,
  pages={4127--4135},
  year={2021}
}

@article{berndt2023receding,
  title={Receding Horizon Re-ordering of Multi-Agent Execution Schedules},
  author={Berndt, Alexander and Van Duijkeren, Niels and Palmieri, Luigi and Kleiner, Alexander and Keviczky, Tam{\'a}s},
  journal={IEEE Transactions on Robotics},
  year={2024},
 volume={40},
  number={},
  pages={1356-1372},
}

@inproceedings{HuangAAAI22,
  author    = {Taoan Huang and Jiaoyang Li and Sven Koenig and Bistra Dilkina},
  title     = {Anytime Multi-Agent Path Finding via Machine Learning-Guided Large Neighborhood Search},
  booktitle = AAAI,
  pages     = {9368--9376},
  year      = {2022}
}

\end{document}